\documentclass[10pt,twocolumn,letterpaper]{article}

\PassOptionsToPackage{dvipsnames,table}{xcolor}

\usepackage{iccv}              

\usepackage{amsmath}
\usepackage{amssymb}
\usepackage{caption}
\usepackage[table]{xcolor}
\definecolor{mygray}{gray}{.95}
\usepackage{marvosym}
\usepackage[accsupp]{axessibility}

\definecolor{iccvblue}{rgb}{0.21,0.49,0.74}
\usepackage[pagebackref,breaklinks,colorlinks,allcolors=iccvblue]{hyperref}

\def\paperID{6446} 
\def\confName{ICCV}
\def\confYear{2025}

\title{RTMap: Real-Time Recursive Mapping with Change Detection and Localization}

\author{Yuheng Du$^{1,\dag}$ \quad Sheng Yang$^{1,\dag,}$\textsuperscript{\Letter} \quad Lingxuan Wang$^1$ \quad Zhenghua Hou$^1$ \quad Chengying Cai$^1$ \\ Zhitao Tan$^1$ \quad Mingxia Chen$^1$ \quad Shi-Sheng Huang$^2$ \quad Qiang Li$^1$ \\
{\normalsize $^1$Unmanned Vehicle Dept., CaiNiao Inc., Alibaba Group \quad
$^2$Beijing Normal University}\\
{\tt\small \{guofan.dyh,shengyang\}@cainiao.com}
}

\begin{document}

\twocolumn[{
\maketitle\centering
\captionsetup{type=figure}
\includegraphics[width=\textwidth]{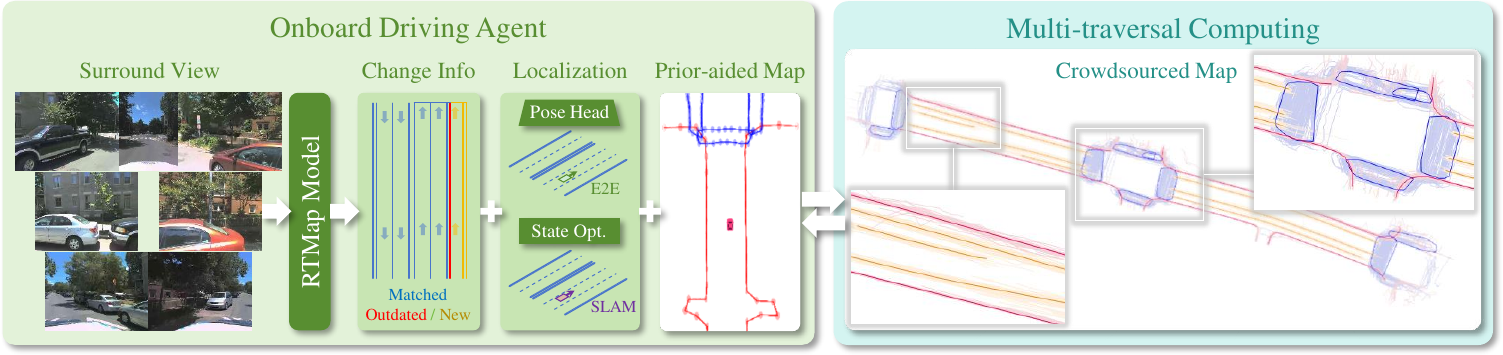}
    \captionof{figure}{RTMap performs real-time online HD mapping on onboard driving agents, simultaneously solving map-based localization, map change detection, and online map fusion tasks. Meanwhile, a cloud service asynchronously stitches crowdsourced online HD maps into a global prior-map for subsequent traverses.}
\label{fig:teaser}
\vspace{5mm}
}]

\let\thefootnote\relax\footnote{$\dag$: Equally contributed. \Letter: Corresponding author.}

\begin{abstract}

While recent online HD mapping methods relieve burdened offline pipelines and solve map freshness, they remain limited by perceptual inaccuracies, occlusion in dense traffic, and an inability to fuse multi-agent observations. We propose RTMap to enhance these single-traversal methods by persistently crowdsourcing a multi-traversal HD map as a self-evolutional memory. On onboard agents, RTMap simultaneously addresses three core challenges in an end-to-end fashion: (1) Uncertainty-aware positional modeling for HD map elements, (2) probabilistic-aware localization w.r.t. the crowdsourced prior-map, and (3) real-time detection for possible road structural changes. Experiments on several public autonomous driving datasets demonstrate our solid performance on both the prior-aided map quality and the localization accuracy, demonstrating our effectiveness of robustly serving downstream prediction and planning modules while gradually improving the accuracy and freshness of the crowdsourced prior-map asynchronously. Our source-code will be made publicly available at \url{https://github.com/CN-ADLab/RTMap}.

\end{abstract}
\section{Introduction}
\label{sec:intro}

High-Definition (HD) maps enriched with geometric, semantic, and behavioral annotations, are foundational to autonomous driving systems -- enabling robust obstacle avoidance, centimeter-level parking precision, and customized driving policies. As highlighted in a recent survey~\cite{wijaya2024high}, the field has shifted from reliance on costly offline mapping pipelines to online HD mapping~\cite{liao2023MapTR,liao2023maptrv2,liu2024mgmap}, which generates maps in real-time using consumer-grade sensors and embedded computing resources. These methods empower vehicles to simultaneously perceive and construct HD maps onboard, unlocking a dual-mode Operational Design Domain (ODD): (1) \emph{First-Traversal Autonomy}: In unexplored environments, vehicles operate without a prior-map, leveraging real-time perception for basic autonomy, e.g., Lane Centering Control (LCC) and Autonomous Emergency Braking (AEB). (2) \emph{Crowdsourced Enhancement}: Once aggregated from multi-agent traversals, such fused prior-maps can provide previous perceptual knowledge to extend sensing range, resolve occlusions, and elevate autonomy to L4 standards, e.g., dense urban navigation and unprotected left turns.

While the community rapidly evolves methods and strategies for the first-traversal autonomy~\cite{wijaya2024high}, \emph{crowdsourced enhancement} remains under-explored following such a shifted architecture. Revisiting online HD mapping frameworks~\cite{liao2023MapTR,liao2023maptrv2,liu2024mgmap}, most of them treat mapping as a single-pass process, failing to effectively utilize the rich contextual knowledge available from multiple traversals of the same scene.
To fully harness this multi-traversal knowledge, two fundamental capabilities are essential: (1) precise localization within the prior-map to accurately align the current surroundings with existing spatial data, and (2) robust detection and adaptation to structural changes in road networks -- such as lane modifications or construction zones -- to maintain map freshness by eliminating outdated information.
Although numerous studies have proposed effective methods to solve these problems independently~\cite{he2023egovm,qin2024crowd}, these two sub-tasks and the overall prior-aided online HD mapping task essentially solve the retrieval, correspondence, and differentiation problems between multi-traversal map elements. To address these challenges, we propose RTMap with novel comprehensive query and matching, to 
solve the element identification and map association in a unified end-to-end framework for addressing all these tasks. Moreover, we further introduce an explicit modeling of the geometric uncertainty of extracted map elements, to facilitate both a probabilistic-aware state estimation of the 6 degrees of freedom (DOF) vehicle poses and noise-aware multi-traversal prior-map fusion. Thus, we lead to highly accurate real-time HD mapping, especially suitable for autonomous driving scenarios.

In summary, our RTMap offers three key contributions:

\begin{itemize}
\item The first end-to-end framework supports multi-traversal online HD mapping while simultaneously addressing map-based localization and change detection to robustly and stably serve downstream autonomous driving modules.
\item By quantitatively inferencing the probabilistic density of perceived vectorized HD map elements, RTMap further improves localization accuracy through end-to-end learning and explicit state estimation.
\item RTMap proposes a crowdsourcing mechanism to recursively update the offline prior-map asynchronously. The inferenced probabilistic density and change information further improve the accuracy of the prior-aided online HD map.
\end{itemize}

Experiments on localization, mapping, and map change detection tasks on multiple publicly available datasets~\cite{lambert2022tbv,caesar2020nuscenes} have demonstrated our effectiveness in simultaneously addressing them through an end-to-end framework. 

\section{Related Work}
\label{sec:rel_work}

In decades, offline mapping pipelines~\cite{yang2018pgo,tang2023thma} have become the mainstream for mass-producing HD maps~\cite{elghazaly2023high}. Once these maps are constructed, centimeter-level map-based localization deployed onboard can accurately fetch and transform these map elements for downstream prediction and planning tasks. However, recent advances~\cite{wijaya2024high} in onboard real-time HD mapping are deeply reforming the solution of mapping (Sec.~\ref{sec:rel_work:mapping}) and localization (Sec.~\ref{sec:rel_work:localization}).

\subsection{Online HD Mapping}
\label{sec:rel_work:mapping}

\noindent\textbf{Single-traversal single moment methods.} HDMapNet~\cite{li2022hdmapnet} and MapTR~\cite{liao2023MapTR,liao2023maptrv2} use the encoder-decoder paradigm to transform sensor input into a unified Bird's Eye View (BEV) representation and output an explicit local HD map. VectorMapNet~\cite{liu2023vectormapnet} aggregates features generated from different modalities into a common BEV feature space for multi-sensor mapping. GKT~\cite{chen2022efficient} leverages calibrated camera parameters to determine the correspondence between 2D positions and BEV grids for constructing full correlations. LaneSegNet~\cite{li2023lanesegnet} proposes a heads-to-regions mechanism for capturing long-range attention and an identical initialization strategy for enhancing the learning of positional priors for lane attention.

\noindent\textbf{Single-traversal temporal fusion methods.} To facilitate the stability of online HD mapping between consecutive frames, many subsequent methods~\cite{yuan2024wacv,li2024dtclmapper} leverage BEVFormer~\cite{li2022bevformer} for temporal self-attention to fuse temporally adjacent BEV information recurrently.
Other explicit methods~\cite{chen2024onlinetemporalfusionvectorized} use odometry to transform perceived results and perform voxel mapping for multi-frame fusion. MapTracker~\cite{chen2024maptracker} leverages BEV and vector memory fusion to better exploit temporal information. PrevPredMap~\cite{peng2025prevpredmap} use previous predictions as priors.

\noindent\textbf{Prior-aided methods.} 
Previous works~\cite{sun2023mind,bateman2024exploring} demonstrate that prior-informed online mapping can further improve completeness and stability.
Unlike the single-traversal methods mentioned above, prior-aided methods require both localization and change detection to establish correct correspondences between maps. 
We mainly found two categories of prior-aided methods for online HD mapping: The first category~\cite{xiong2023neuralprior,wu2023mapnerf} uses neural representation as prior-maps, which is storage-intensive and not suitable for large-scale deployment or manual inspection. The second category uses Standard-Definition (SD) or HD maps as prior knowledge: SMERF~\cite{luo2024augmenting} queries the SD map at the ego vehicle's location for lane-topology reasoning, and U-BEV~\cite{camiletto2024ubev} extracts BEV features from both sensors and SD map to reach meters-level accuracy through template matching. P-MapNet~\cite{jiang2024pMapNet} leverages SD priors for coarse localization and HD priors for final map refinement. PriorDrive~\cite{zeng2024driving} integrates diverse prior maps through the Unified Vector Encoder. Our method belongs to the latter category, stepping further to reach real-time centimeter-level localization accuracy beyond previous methods through an end-to-end model. Meanwhile, RTMap manages to self-enclose the maintenance of prior HD maps, inferencing the probabilistic density of online HD map elements for noise-aware HD crowdsourcing.

\begin{figure*}[t]
\centering
\includegraphics[width=\linewidth]{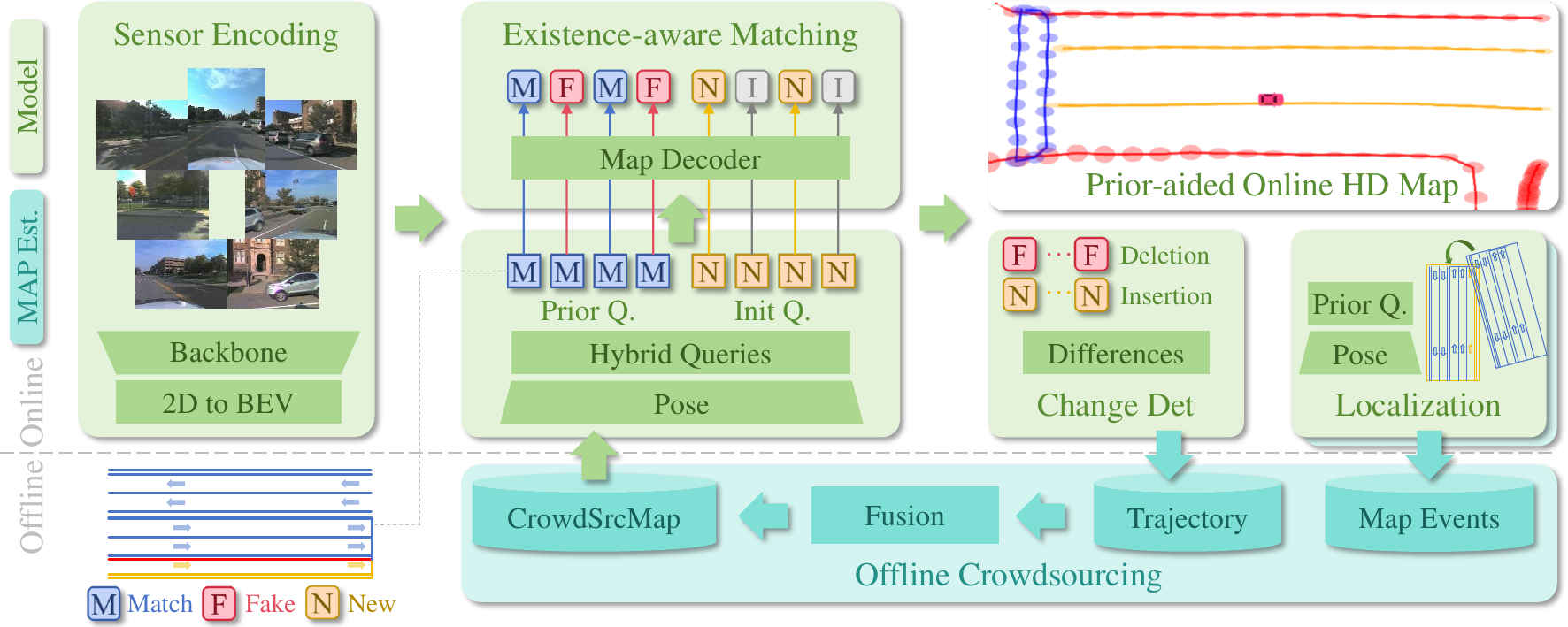}
\caption{Online and offline modules of RTMap. We encode sensors and the crowdsourced HD map to perform hybrid queries and existence-aware matching to obtain matched, outdated, and newly observed HD map elements online. Matched queries can be used for either a pose head or a maximum a posteriori (MAP) state estimator for obtaining the 6-DOF vehicle poses. During offline, we gather multi-traversal local HD maps to fuse them to improve the accuracy and completeness of the crowdsourced map.}
\label{fig:pipeline}
\end{figure*}

\noindent\textbf{Change detection.}
Considering possible road structural changes, recent methods such as ExelMap~\cite{wild2024exelmap} add additional heads for detecting element-wise insertion and deletion events. Also, recent advances for offline pipelines~\cite{xia2025ldmapnet} leverage prior-map encoding and unify the feature space for the predicted and historical instances using distinct multi-layer perceptron (MLP) networks to associate and pick up changed instances. Compared to these methods, our method operates onboard and thus instantly serves downstream tasks without relying on engineering practices to ensure the refreshed HD map is efficiently deployed.

\subsection{Map-based Localization}
\label{sec:rel_work:localization}

Online HD mapping reforms localization separately into odometry~\cite{zhao2024viwo,lee2024lidar} for temporal reasoning and map-based localization~\cite{cheng2021roadmapping,miao2024survey} for aggregating prior knowledge if such prior exists. We found several learning-based methods contain similar architectures above but are individually designed for map positioning tasks: BEV-Locator~\cite{zhang2022bev} adopts the transformer structure in cross-modal feature association to address cross-modality matching between semantic map elements and camera images. EgoVM~\cite{he2023egovm} designs learnable embeddings and a transformer decoder to bridge the representation gap between vectorized HD maps and sensor BEV features to reach centimeter-level localization accuracy. However, these methods treat localization as an isolated task and still require a standalone HD mapping progress for referring to. Considering the main objective of a precise map-based localization -- better aggregating prior knowledge for map servicing, we choose to union the localization task with all related tasks that may influence its accuracy -- online map prediction accuracy, road structural changes, erroneous map elements matching, and possible occlusions -- and thus lean toward a multi-task end-to-end manner.

Meanwhile, inspired by Gu et al.~\cite{GuSongEtAl2024}, who additionally output uncertainty estimates for downstream prediction tasks, we contribute to introduce uncertainty as thorough information, which can either explicitly or implicitly improve the quality of localization and prior-aided mapping, enabling end-to-end architecture to optionally perform optimization-based state estimation as Simultaneous Localization and Mapping (SLAM) approaches~\cite{bowman2017em,cao2018realtime,huang2020lidar,du2020accurate,zou2024gp} for interpretability while maintaining accuracy.

\section{Methology}
\label{sec:method}

\subsection{Method Overview}
\label{sec:method:problem}

\noindent\textbf{Problem formulation.} In the current traverse $t$ given a single moment of multi-sensor frames $\mathbf{I}_t$ (typically contains calibrated surround view camera frames) and a prior-map $\mathcal{M}_{t-1}$ formed by previous multi-agent traverses, our end-to-end framework simultaneously addresses the following tasks:
\begin{equation}
\begin{aligned}
\begin{aligned}
\{\mathbf{M}_t,\mathbf{U}_t,\mathbf{D}_t ,\mathbf{T}_t^\mathbb{E} \} &\leftarrow \mathrm{RTMapModel}(\mathbf{I}_t, \mathcal{M}_{t-1}), \\
\{\mathbf{T}_t^\mathbb{R}\} &\leftarrow \mathrm{Localize}(\mathbf{M}_t,\mathbf{U}_t, \mathbf{D}_t,\mathcal{M}_{t-1}), \\
\{\mathcal{M}_t\} &\leftarrow\mathrm{CSrc}(\mathbf{M}_{t},\mathbf{U}_t,\mathbf{D}_t,\mathbf{T}_t,\mathcal{M}_{t-1}). \\
\end{aligned} \\
\end{aligned}
\end{equation}

As illustrated in Fig.~\ref{fig:pipeline}, we use a multi-task onboard model $\mathrm{RTMapModel(\cdot)}$ with hybrid queries (Sec.~\ref{sec:method:hybrid_query}) and an existence-aware matching scheme (Sec.~\ref{sec:method:matching}) to detect three classes of map elements: matched $\mathbf{M}_t^\mathbb{M}$, outdated $\mathbf{M}_t^\mathbb{F}$, and newly observed $\mathbf{M}_t^\mathbb{N}$.
Therefore, we can separately infer possible map change events $\Delta_t \triangleq \{\mathbf{M}_t^{\mathbb{F}}, \mathbf{M}_t^{\mathbb{N}} \}$ besides matched elements $\mathbf{M}_t^\mathbb{M}$ to reduce their impact on localization and update the crowdsourced map afterward. 
We design our loss function (Sec.~\ref{sec:method:loss}) for multiple additional tasks, jointly inferencing (1) the per-vertex probabilistic density $\mathbf{U}_t$ of these map elements, (2) point-wise correspondences $\mathbf{D}_t$ between re-observed and prior-map elements $\left\langle \mathbf{M}_t^\mathbb{M}, \mathcal{M}_{t-1} \right\rangle$, and (3) an end-to-end pose w.r.t. the prior-map $\mathbf{T}_t^{\mathbb{E}} \in \mathbb{SE}^3$.
For explicitly addressing the 6-DOF localization if required, we also implement a state optimizer $\mathrm{Localize(\cdot)}$ leveraging deducted correspondences $\mathbf{D}_t$ for an optimization-based pose solution $\mathbf{T}_t^{\mathbb{R}} \in \mathbb{SE}^3$ as an option (Sec.~\ref{sec:method:slam}). As $\mathbf{T}_t^{\mathbb{E}}$ and $\mathbf{T}_t^{\mathbb{R}}$ share the same purpose, we compare their performance (Tab.~\ref{tab:localization_nuscene}) and retain the estimator of $\mathbf{T}_t^{\mathbb{R}}$ for tightly coupling inertial measurements if necessary. Finally, $\mathrm{CSrc}(\cdot)$ operates on-the-cloud to fuse multi-traversal inferences and update the crowdsourced map (Sec.~\ref{sec:method:slam}). We use GPS measurements to provide a meters-level initial pose $\mathbf{T}_t^{0}$ for fetching a truncated local crowdsourced HD map $\mathcal{M}_{t-1}$ for these onboard operations.

\subsection{Hybrid Queries}
\label{sec:method:hybrid_query}

To tackle the challenges of localization, mapping, and change detection concurrently, we propose hybrid queries enabling the network to process various query types and obtain matched, outdated, and newly observed elements to address all three tasks effectively. In this context, the decoder initially receives two types of queries $\mathbf{Q}_{\mathrm{prior}}$ and $\mathbf{Q}_{\mathrm{new}}$ from prior-map embedding and sensor encoding, respectively. After training with query-based matching (Sec.~\ref{sec:method:matching}), we further differentiate $\mathbf{Q}_\mathrm{prior} \rightarrow \{ \mathbf{Q}_\mathrm{map}, \mathbf{Q}_\mathrm{fake}\}$, with the first part $\mathbf{Q}_\mathrm{map}$ telling reliable correspondences for localization and the latter part $\mathbf{Q}_\mathrm{fake}$ specifying outdated elements. 

Inspired by MapTR~\cite{liao2023MapTR} and MapEX~\cite{sun2023mind}, we use unified representation to encode each $\mathbf{Q}_{\mathrm{prior}}$ as an instance represented by fixed points, where the first two dimensions of each point are encoded as $\mathrm{XOY}$ coordinates, and the remaining $\mathrm{N}$ dimensions represent category information using one-hot encoding. Due to the characteristics of the transformer, the number of queries is fixed. Therefore, the remaining queries are designated as $\mathbf{Q}_{\mathrm{new}}$ padded with zeros. Finally, after passing through a linear layer, these queries are combined with the hierarchical query embeddings $\mathbf{Q}_{\mathrm{hie}}$~\cite{liao2023MapTR} to obtain the hybrid queries $\mathbf{Q}_{\mathrm{hybrid}}$ formulated as follows: 
\begin{equation}
\mathbf{Q}_{\mathrm{hybrid}} = \{\mathbf{Q}_{\mathrm{map}}, \mathbf{Q}_{\mathrm{fake}},\mathbf{Q}_{\mathrm{new}}\} + \mathbf{Q}_{\mathrm{hie}}.
\label{equ:queries}
\end{equation}

\subsection{Existence-aware Matching}
\label{sec:method:matching}



\noindent\textbf{Matching in training}.
In contrast to MapEX~\cite{sun2023mind}, which leverages pre-attribution from ground truth for training, we need to tell $\mathbf{Q}_\mathrm{fake}$ apart from $\mathbf{Q}_\mathrm{map}$ for change detection and improving multi-task performances. Therefore, with the help of ground truth map-changing events, we choose to implement only a pre-attribution of $\mathbf{Q}_{\mathrm{map}}$ predictions to their corresponding existing map elements. Besides, the remaining map elements are matched with $\mathbf{Q}_{\mathrm{new}}$ predictions using conventional Hungarian matching.

\begin{figure}
\includegraphics[width=\linewidth]{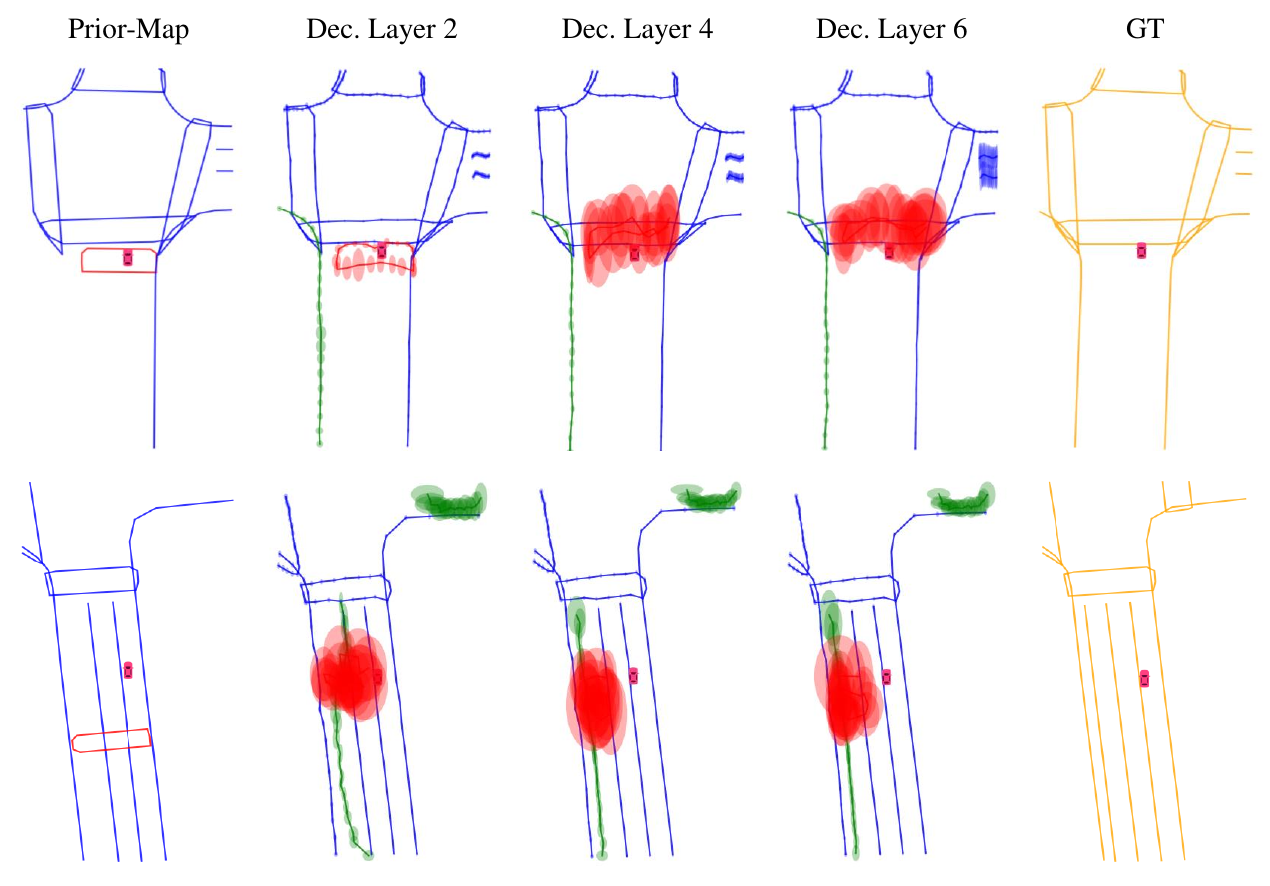}
\caption{Trend of queries in different decoder layers. $\mathbf{Q}_{\mathrm{map}}$, $\mathbf{Q}_{\mathrm{fake}}$, $\mathbf{Q}_{\mathrm{new}}$ are colorized in blue, red, and green, respectively. The ellipsoid reflects their uncertainty. Hence, outdated map elements are gradually filtered from matched map elements.}
\label{fig:query}
\end{figure}

As an observable consequence of the training process (Fig.~\ref{fig:query}), different types of queries exhibit distinct behaviors under this matching strategy. Due to pose perturbations, Map queries $\mathbf{Q}_{\mathrm{map}}$ are projected near their actual locations. After that, their corresponding reference points gradually move to the correct positions. In contrast, since fake queries $\mathbf{Q}_{\mathrm{fake}}$ do not exist in the prior-map, they lead to unstable behavior in their reference points. The rest of the new queries $\mathbf{Q}_{\mathrm{new}}$ detect newly occurring elements from the current traverse. 

\noindent\textbf{Matching in inference}.
During the inference stage, without ground truth annotations, $\mathbf{Q}_{\mathrm{map}}$ and $\mathbf{Q}_{\mathrm{fake}}$ are mixed within the prior queries $\mathbf{Q}_{\mathrm{prior}}$. Since $\mathbf{Q}_{\mathrm{fake}}$ were not pre-attributed during the training phase, the category confidence associated with $\mathbf{Q}_{\mathrm{fake}}$ will be significantly lower than those for $\mathbf{Q}_{\mathrm{map}}$, providing strong evidence to differentiate between the two types of queries. Therefore, we believe that using the confidence score of the predictions corresponding to $\mathbf{Q}_{\mathrm{prior}}$ can effectively distinguish between them.

\subsection{Loss Function for Map Elements and Poses}
\label{sec:method:loss}

\noindent\textbf{Composite Geometric Loss for Element Vertices.}
Following Gu et al.~\cite{GuSongEtAl2024}, we choose 
univariate Laplace distributions
to model map elements, allowing us to additionally output the 
uncertainty of the location $\mathbf{M}_t$ of map element vertices.
For inferencing such additional information, we augment a negative log-likelihood (NLL) loss $\mathbf{L}_{\mathrm{nll}}$ for each vertex, as:
\begin{equation}
\begin{aligned}
\mathbf{L}_\mathrm{pts}&=\lambda_1 \cdot \mathbf{L}_\mathrm{nll} + \lambda_2 \cdot \mathbf{L}_\mathrm{mht},  \\
\mathbf{L}_\mathrm{nll} &= \sum_{v=1}^\mathrm{V}\sum_{k=1}^{2}\left(\log \left(2 \sigma_{v}^{k}\right)+\frac{|\mathbf{m}_{v}^{k}-\mu_{v}^{k}|}{\sigma_{v}^{k}} \right), \\
\end{aligned}
\label{equ:nll}
\end{equation}
We keep their $\mathrm{Z}$-value representation unchanged due to insufficient observability. The parameters $\mu_v^{k},\sigma_v^k \in \mathbb{R}$ denote the location and scale parameters of the $k^\mathrm{th}$ dimensional 
Laplace distributions
for a vertex $\mathbf{m}_v$, and we use a fixed total size $\mathrm{V}=20$ for each map element. We remain the original manhattan distance loss $\mathbf{L}_\mathrm{mht}$~\cite{liao2023maptrv2} with hyper-parameters $\lambda_1$ and $\lambda_2$ to balance the convergence ability and the element accuracy.

\noindent\textbf{Pose Auxiliary Loss}.
Given a coarse initial pose $\mathbf{T}^0$, the $\mathbf{Q}_{\mathrm{map}}$ of the input prior-map may be incorrectly positioned. Hence, the pose auxiliary loss $\mathbf{L}_\mathrm{pose}$ is designed to provide additional supervision for the transformer decoder, helping it learn to correct such misalignment more effectively. Our end-to-end design utilizes the output features from the decoder to precisely align perceived map elements with the prior-map. Regarding map alignment, information can be retrieved with in the map queries $\mathbf{Q}_{\mathrm{map}}$, which are then added to the output of each decoder layer. Subsequently, we utilize a shared MLP and a max-pooling layer to obtain the predicted delta pose $\partial\hat{\mathbf{T}}_\mathrm{aux}$. Finally, the predicted delta pose is supervised by ground truth delta pose $\partial \mathbf{T}$ with a smooth L1 Loss.

\subsection{MAP for Localization and Crowdsourcing}
\label{sec:method:slam}

\noindent\textbf{Optimization-based Localization.} For the $\mathrm{Localize(\cdot)}$ operation, we can explicitly remove map-changing events $\Delta_t \triangleq \{\mathbf{M}_t^{\mathbb{F}}, \mathbf{M}_t^{\mathbb{N}} \}$ from the matched associations $\mathbf{D}_t:\left\langle \mathbf{M}_t^\mathbb{M}, \mathcal{M}_{t-1} \right\rangle$ for outlier rejection, and employ the additional probabilistic density of vertices $\mathbf{U}_t$. Specifically, we launch the following maximum a posteriori (MAP) state estimation for solving the vehicle pose: 
\begin{equation}
\begin{aligned}
&\min_{\mathbf{T}^\mathbb{R}} \mathbf{E}^1(\mathbf{D}_t), \\
\mathbf{f}^1 \propto &\exp(-\frac{1}{2} \left\| \mathbf{T}^\mathbb{R} \cdot \mathbf{m}_{t}^i - m_{t-1}^i \right\|^2_{\mathbf{g}_{t}^i}), \\
\end{aligned}
\label{equ:map}
\end{equation}
where $\mathbf{E}^1(\mathbf{D}_t)=\sum_{\mathbf{D}_t} -\log(\mathbf{f}^1)$ is the sum of the negative log-likelihood of point-to-point residuals $\mathbf{f}^1$, and we use the squared Mahalanobis distance whose covariance is assigned as the Gaussian mixture $\mathbf{g}_{t}^i \triangleq \mathbf{u}_{t}^i\oplus u_{t-1}^i$ of both inferenced and crowdsourced probabilistic densities~\cite{cao2018realtime}. Each pair of associations denoted as $d_t^i:\left\langle \mathbf{m}_{t}^i, m_{t-1}^i \right\rangle \in \mathbf{D}_t$ constructs a residual between corresponded vertices across two maps. One can combine more residuals like IMU pre-integration~\cite{forster2016manifold} or odometry~\cite{qin2018vins,zhao2024viwo} for a tightly coupled system, and in this paper, we mainly test the difference between the single residual form $\mathbf{E}^1$ and the end-to-end regressed pose $\mathbf{T}^\mathbb{E}$. 
Specifically, we use the Levenberg-Marquardt algorithm to minimize $\mathbf{E}^1(\mathbf{D}_t)$.

\noindent\textbf{Probabilistic-aware Crowdsourcing.} For the $\mathrm{CSrc}(\cdot)$ operated on-the-cloud, we gather multi-traversal and multi-frame observations $\langle \mathbf{M}_{t}^{j}, \mathbf{U}_{t}^{j}, \mathbf{D}_{t}^{j} \rangle$ with their relative pose w.r.t. a base frame as $\mathbf{T}_{t}^{j}$, to fuse the crowdsourced map again in a MAP manner. For each tracked map vertex with multiple observations, we use the union-find algorithm to construct the following position solver for its latest position $m_t \in \mathcal{M}_t$:
\begin{equation}
\min_{m_t} \sum_{\hat{\mathbf{m}}_t^j \in \mathbf{M}} \frac{1}{2} \| m_t - \mathbf{T}_t^j \cdot \hat{\mathbf{m}}_t^j \|_{\mathbf{u}_t^j}^2 + \frac{1}{2}\| m_t - m_{t-1}\|^2_{u_{t-1}},
\label{equ:fuse}
\end{equation}
where $\hat{\mathbf{m}}_t^j$ represents an associated observation $\mathbf{M}_{t}^{j}$, and the union $\mathbf{M}$ gathers all matched observations, and we again use the Gaussian mixture model to continuously refine the probabilistic density $u_t$ for each vertex. Meanwhile, we perform Hungarian voting to determine the predecessor and successor vertices of each vertex, to preserve the topology of HD map elements. For the first traverse, the second term leveraging the previous crowdsourced version is omitted for an initial crowdsourcing.

\section{Implementation, Results, and Evaluation}
\label{sec:eval}

\subsection{Implementation details}
\label{sec:eval:impl}

\noindent\textbf{Training Sample Generation.} 
To better align with real-world scenarios and  
enhance network robustness,
we first apply random perturbations to the ground truth pose $\mathbf{T}^{\mathrm{gt}}$ to generate a noisy initial pose estimate $\mathbf{T}^{\mathrm{init}}$. 
Using
$\mathbf{T}^{\mathrm{init}}$, we then crop the ground truth map. Occasionally, when the local map obtained from the noisy pose is the same size as the perceptual range, some map elements may fall outside the perceptible area. 
To ensure all ground truth map elements remain accessible, we pad the perceptual range as necessary.
Next,
we randomly remove certain elements from the prior-map and generate synthetic fake map elements. We do not perturb the ground truth map elements to simulate specific changes (such as shifting lane markings), as such modifications could be interpreted as deleting the original map elements and adding new ones.


\noindent\textbf{Implementation and Training Details.} We choose ResNet50~\cite{he2016resnet} as the image backbone. Our RTMap model is trained using 8 NVIDIA GeForce RTX 3090 GPUs with a total batch size of 32 for 36 epochs. We utilize the AdamW optimizer with a learning rate of $6\times10^{-4}$. To ensure fair comparisons, we adopt the same training settings for MapTRv2~\cite{liao2023maptrv2} and ensure its convergence. For $\mathbf{L}_\mathrm{pts}$ in Eq.~\ref{equ:nll}, we set \(\lambda_1 = 0.03\) and \(\lambda_2 = 5.0\) for balancing their influence. For $\mathbf{L}_\mathrm{pose}$, the translation weight is set to \(0.04\), corresponding to the scale in radians used when predicting the heading angle.
We set the longitudinal range to $[-36, 36]$ and the lateral range to $[-18, 18]$ meters for online HD mapping.
In contrast to prior-aided method such as HRMapNet~\cite{zhang2024enhancing}, we do not use crowdsourced prior-maps for training. Instead, we use artificially perturbed prior-maps to ensure we have incorporated sufficient map-changing events regarding the current scale and maturity of publicly available datasets.


\subsection{Experiment Setup}
\label{sec:eval:exp}


\noindent\textbf{Datasets.} We evaluate the performance of the proposed network using two datasets: TbV~\cite{lambert2022tbv} and nuScenes~\cite{caesar2020nuscenes}. The TbV dataset~\cite{lambert2022tbv} provides over 200 scenarios involving real-world changes. These map-changing events are primarily on lane topology, road boundaries, and pedestrian crossings. It is specifically designed to detect discrepancies between sensor data and HD maps caused by these changes, with the val set containing scenarios that feature actual map alterations. For training, the dataset also incorporates synthetically modified ground truth maps. The nuScenes dataset~\cite{caesar2020nuscenes} is a large-scale dataset for autonomous driving, consisting of 1,000 driving scenes in urban environments, each sampled at 2Hz. It includes sensor data such as RGB images, LiDAR point clouds, inertial measurements, and labeled HD maps for evaluation.

\noindent\textbf{Tasks and Metrics.} Regarding the insufficiency of multi-traversal scanning and map-changing events, 
we tested the main task -- crowdsourcing -- as well as two sub-tasks, change detection and localization, on the TbV dataset.
Meanwhile, we use the nuScenes dataset to test the performance of our map-based localization w.r.t. its labeled HD map. 

Unlike conventional online HD mapping tasks, our crowdsourcing task accounts for localization errors. Therefore, we introduce localization perturbations on top of the evaluation method used in online HD mapping tasks. Specifically, for a more realistic simulation of localization noise, we sample perturbations from Gaussian distributions: a lateral perturbation of $\mathcal{N}(0, 0.75^2)$, a longitudinal perturbation of $\mathcal{N}(0, 1.5^2)$ in meters, and a yaw perturbation of $\mathcal{N}(0, 0.85^2)$ in degrees.
To obtain ground truth for evaluation, we augment the map data by aligning the previous validation set with sensor data for ensuring 3D consistency. Specifically, we use average precision to evaluate the quality of the generated map and Chamfer distance to determine the alignment between the ground truth and predicted maps. We compute the AP at the following thresholds \( \{0.5, 1.0, 1.5\} \) in meters keeping the same with experiments in other methods~\cite{liao2023maptrv2}, and then calculate the mean to obtain the final mean average precision ($\mathrm{mAP}$).

For the change detection task, we follow the evaluation implementation of the officially released code in TbV~\cite{lambert2022tbv}, to treat the change detection task as a binary classification problem ($\mathrm{Acc}$), with classes indicating whether a change has occurred $\mathrm{Acc}_{c}$ or not $\mathrm{Acc}_r$, and the mean accuracy (\(\mathrm{mAcc}\)) is also calculated to assess overall performance.

For the localization task, we measure the mean and 90\textsuperscript{th} absolute error of ego-poses relative to the ground truth HD map in three main dimensions: lateral, longitudinal, and yaw.

\subsection{Performance on Crowdsourcing}
\label{sec:eval:cs}

\begin{table}[t]
\caption{Quantitative comparison of different online mapping approaches and strategies on TbV~\cite{lambert2022tbv}. We additionally perform an ablation study on whether or not to use the vertex-level probabilistic density for crowdsourcing.}
\centering
\begin{tabular}{l|c|ccc|c}
\toprule
\toprule
  &   & \multicolumn{4}{c}{$\mathrm{mAP(\%)}\uparrow$} \\
Method & Cycle & Ped. & Div. & Bou. & Avg. \\
\midrule
\midrule
\rowcolor{mygray}\multicolumn{6}{c}{Straight} \\
MapTRv2 & Ave. & 31.7 & 42.0 & 37.3 & 37.1 \\
HRMapNet & Ave. & 34.2 & 43.7 & 39.8 & 39.2 \\
MapTracker & Ave. & 35.7 & 44.6 & 39.6 & 39.9 \\
Ours (w/o $\mathbf{U}$) & 2 & 28.6 & 60.5 & 31.7 & 40.2 \\
Ours    & 2 & 32.7 & 68.6 & 35.5 & 45.6 \\
Ours (w/o $\mathbf{U}$) & 3 & 35.7 & 74.4 & 42.0 & 50.7 \\
Ours    & 3 & \textbf{40.9} & \textbf{84.3} & \textbf{47.6} & \textbf{57.6} \\
\midrule
\rowcolor{mygray}\multicolumn{6}{c}{Turning} \\
MapTRv2 & Ave. & 28.2   & 31.6   & 18.3   & 26.0   \\
HRMapNet & Ave. & 31.0 & 32.8 & 19.6 & 27.8 \\
MapTracker & Ave. & 31.4 & 33.2 & 19.3 & 28.0 \\
Ours (w/o $\mathbf{U}$) & 2 & 30.2   & 65.5   & 30.1   & 41.9   \\
Ours    & 2 & 33.1   & 71.0   & 32.3   & 45.5   \\
Ours (w/o $\mathbf{U}$) & 3 & 35.9   & 71.2   & 32.6   & 46.5   \\
Ours    & 3 & \textbf{42.3}   & \textbf{85.2}   & \textbf{38.8}   & \textbf{55.4}   \\
\bottomrule
\bottomrule
\end{tabular}
\label{tab:csrc_tbv}
\end{table}

\begin{figure*}[t]
\centering
\includegraphics[width=0.95\linewidth]{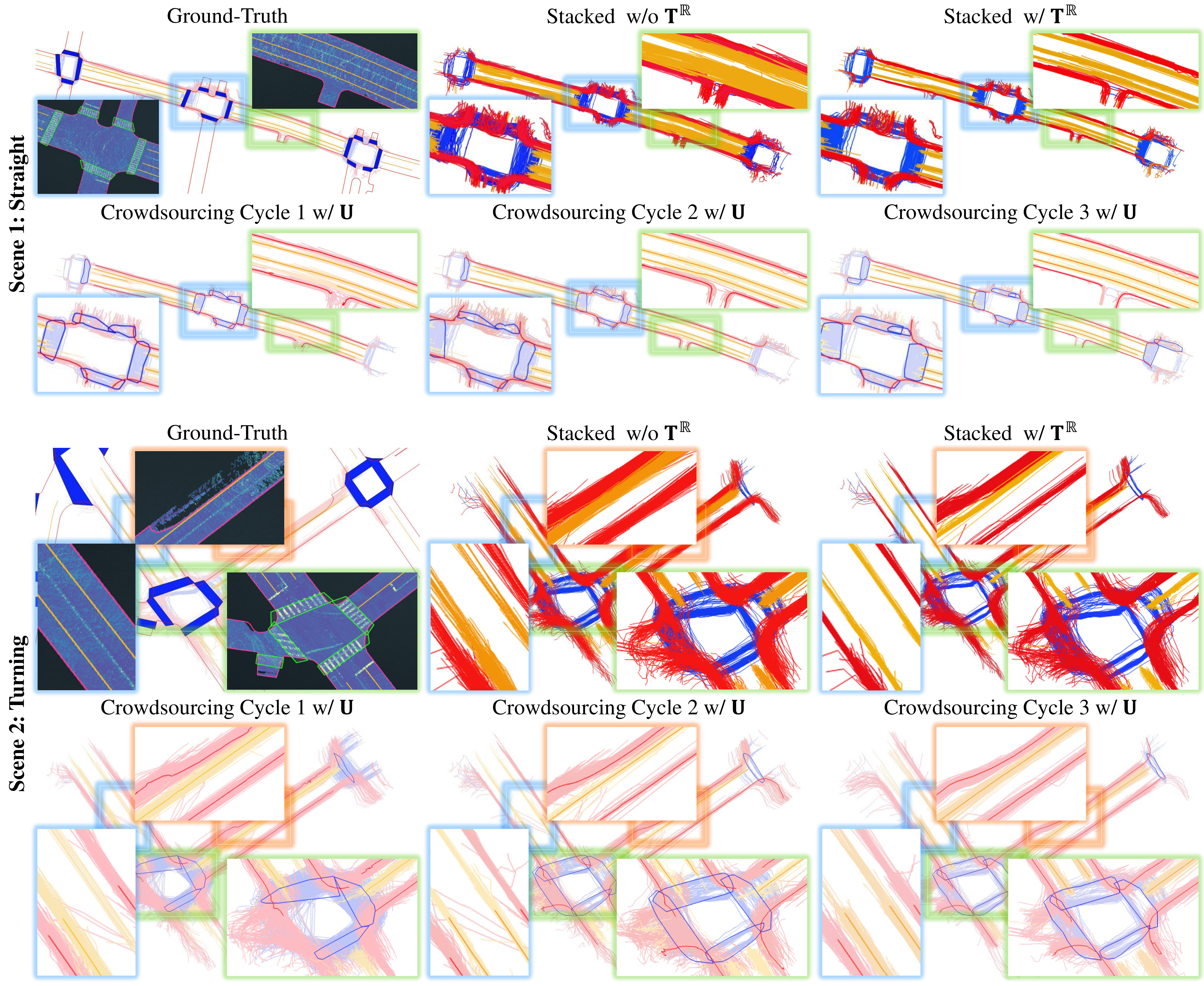}
\vspace{-3pt}
\caption{Qualitative comparison trending the quality improvement through solving a precise pose for alignment, and a multi-traversal probabilistic-aware crowdsourced HD mapping. We zoom-in several map regions for further comparisons.}
\label{fig:crowdsource}
\end{figure*}

We cluster multi-traversal spatially overlapped driving clips in TbV for performing crowdsourcing, group 15 found clips into two scenarios -- straight (6) and turning (9), and listed our results 
in comparison with state-of-the-art online HD mapping methods,
as shown in Tab.~\ref{tab:csrc_tbv}. 
Due to the introduction of localization noise, the accuracy of existing online HD mapping methods is significantly reduced. Our approach is essentially the same as existing methods in the first cycle, when neither localization nor change detection is required. In later cycles, by utilizing the prior-map, our method gradually decreases localization errors and improves mapping accuracy.
We also compare to a direct version, which does not infer and use probabilistic densities $\mathbf{U}$ during crowdsourcing, highlighting the effectiveness of leveraging such uncertainty in mapping tasks. We also refer readers to Fig.~\ref{fig:crowdsource}, which presents the trend of quality improvement by solving a precise trajectory and experiencing multi-traversals.

\subsection{Performance on Change Detection}
\label{sec:eval:changedet}
To evaluate the effectiveness of the proposed RTMap under dynamic driving conditions, we performed change detection tests on the TbV dataset, with modifications including `insert crosswalk' and `delete crosswalk'. 

The results of our experiments, presented in Tab.~\ref{tab:changedet_acc}, indicate a clear distinction in performance across the different scenarios, with our model demonstrating a particularly high sensitivity to the `changed' category. This suggests that RTMap is well-suited to recalling possible map-changing events. Practically, a higher recall enables crowd-mining on these dubious events and ensures safety. Nevertheless, our method achieves higher overall accuracy than the previous method.

From the results, we can conclude that the RTMap’s performance is robust in dynamic contexts. This capability not only enhances the model's effectiveness but also indicates its potential applicability in real-world scenarios, where rapid adaptation to environmental changes is essential.

\begin{table}[htb]
\caption{Quantitative comparison of change detection between ours and the original approach proposed for TbV~\cite{lambert2022tbv}.}
\centering
\begin{tabular}{l|cc|c}
\toprule
\toprule
Method & $\mathrm{Acc_{c}(\%)}\uparrow$ & $\mathrm{Acc_{r}(\%)}\uparrow$ & $\mathrm{mAcc(\%)}\uparrow$ \\
\midrule
\midrule
TbV~\cite{lambert2022tbv} & 40.0 & \textbf{68.2} & 54.1 \\
RTMap  & \textbf{48.9}    & 66.0 & \textbf{57.4}    \\
\bottomrule
\bottomrule
\end{tabular}
\label{tab:changedet_acc}
\end{table}

\vspace{-2pt}

\subsection{Performance on Localization}
\label{sec:eval:localization}

\begin{figure}[t]
\includegraphics[width=\linewidth]{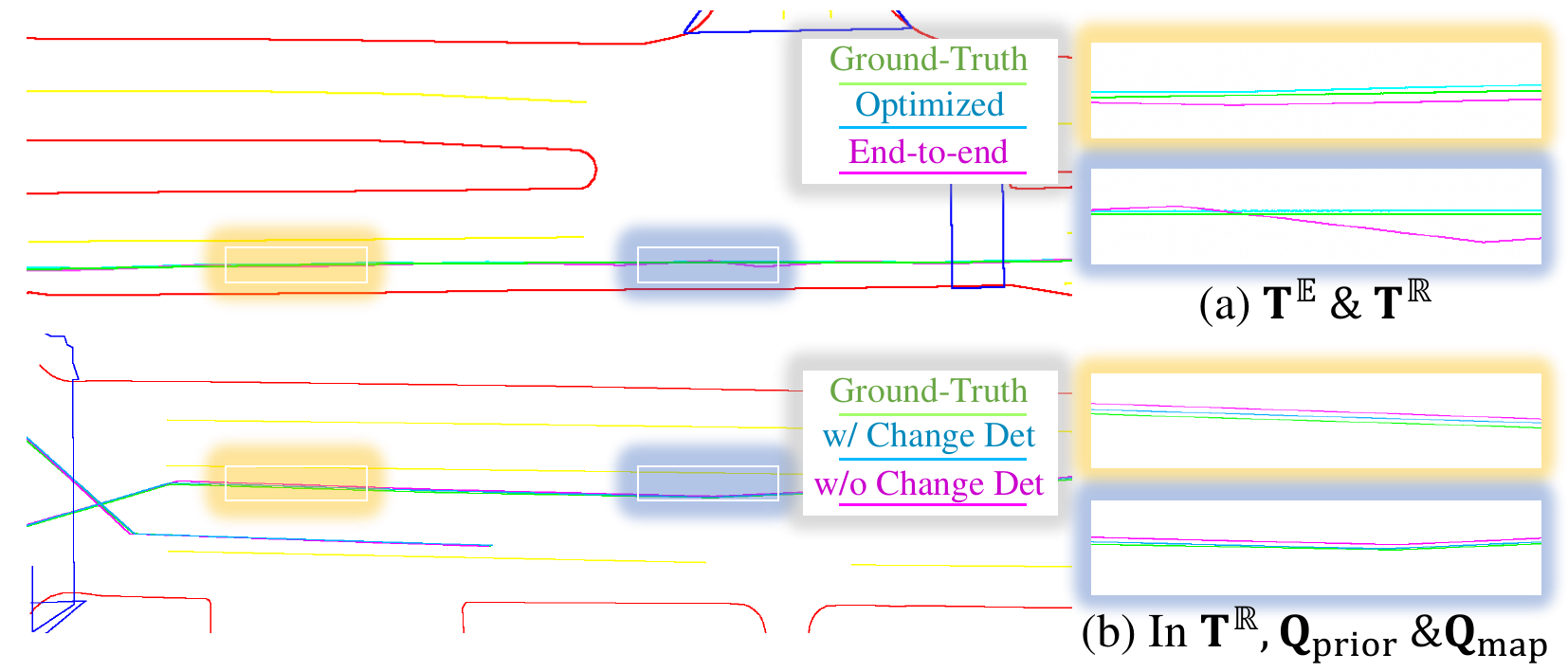}
\caption{Qualitative comparison of the localization performance between two configurations on TbV~\cite{lambert2022tbv}, we show deviations of our estimated trajectory w.r.t. the ground truth, to illustrate the effectiveness of leveraging state optimization ($\mathbf{T}^\mathbb{R}$) and map changing events ($\mathbf{Q}_\mathrm{map}$) for further improving the accuracy.}
\label{fig:loc_comp}
\end{figure}

Our end-to-end framework also supports localizing the vehicle w.r.t. the prior-map. In the first experiment, we perform an ablation study on whether or not to incorporate the change detection task -- differentiating $\mathbf{Q}_\mathrm{map}$ from $\mathbf{Q}_\mathrm{prior}$, and show the results in Tab.~\ref{tab:localization_tbv} and Fig.~\ref{fig:loc_comp}-(b), which proves the advantage of tightly coupling the change detection task to reject mismatched map elements for localization.

\begin{table}[htb]
\caption{An ablation on whether or not using hybrid queries to distinguish $\mathbf{Q}_\mathrm{fake}$ from $\mathbf{Q}_\mathrm{map}$ on TbV~\cite{lambert2022tbv}.}
\centering
\setlength{\tabcolsep}{2.5pt}
\begin{tabular}{l|cc|cc|cc}
\toprule
\toprule
 & \multicolumn{2}{c|}{Lat (m)$\downarrow$} & \multicolumn{2}{c|}{Lon (m)$\downarrow$} & \multicolumn{2}{c}{Yaw ($^\circ$)$\downarrow$} \\
Method & Mean & 90\textsuperscript{th} & Mean & 90\textsuperscript{th} & Mean & 90\textsuperscript{th} \\
\midrule
\midrule
RTMap ($\mathbf{Q}_\mathrm{prior}$) & 0.163 & 0.318 & 0.686 & 1.616 & 0.332 & 0.766 \\
RTMap ($\mathbf{Q}_\mathrm{map}$) & \textbf{0.125} & \textbf{0.256} & \textbf{0.633} & \textbf{1.530} & \textbf{0.317} & \textbf{0.713} \\
\bottomrule
\bottomrule
\end{tabular}
\label{tab:localization_tbv}
\end{table}

We also evaluate localization on the nuScenes dataset as listed in Tab.~\ref{tab:localization_nuscene}, which includes 100 different scenarios from the validation set. We find the following conclusions through experimental results: (1) The explicit optimization-based pose estimation $\mathbf{T}^\mathbb{R}$ still outperforms the end-to-end manner $\mathbf{T}^\mathbb{E}$, (2) the proposed loss functions in Sec.~\ref{sec:method:loss}, including the localization loss $\mathbf{L}_\mathrm{pose}$ and the probabilistic density aware vertex regression loss $\mathbf{L}_\mathrm{pts}$ during training, can effectively contribute to the localization task.

In the current framework, we no longer require a standalone map-based localization task~\cite{zhang2022bev,he2023egovm} to serve downstream planning and control modules. Instead, we concentrate on tightly coupling change detection, localization, and the prior-map together, to benefit the completeness and robustness of the local map for serving downstream planning and control modules.

\begin{table}[htb]
\caption{Ablations of accuracy between end-to-end pose estimation $\mathbf{T}^\mathbb{E}$, and optimization-based pose estimation $\mathbf{T}^\mathbb{R}$ under different settings on nuScenes~\cite{caesar2020nuscenes}.}
\centering
\setlength{\tabcolsep}{3.3pt}
\begin{tabular}{l|cc|cc|cc}
\toprule
\toprule
 & \multicolumn{2}{c|}{Lat (m)$\downarrow$} & \multicolumn{2}{c|}{Lon (m)$\downarrow$} & \multicolumn{2}{c}{Yaw ($^\circ$)$\downarrow$} \\
Method & Mean & 90\textsuperscript{th} & Mean & 90\textsuperscript{th} & Mean & 90\textsuperscript{th} \\
\midrule
\midrule
RTMap ($\mathbf{T}^\mathbb{E}$) & 0.142 & 0.306 & 0.589 & 1.447 & 0.521 & 1.091 \\
RTMap ($\mathbf{T}^\mathbb{R}$) & 0.121 & 0.257 & \textbf{0.586} & \textbf{1.429} & \textbf{0.368} & \textbf{0.799} \\
\midrule
$\mathbf{T}^\mathbb{R}$: Base & 0.122 & 0.254 & 0.618 & 1.510 & 0.390 & 0.840 \\
$\mathbf{T}^\mathbb{R}$: +$\mathbf{L}_\mathrm{pose}$ & 0.122 & 0.261 & 0.590 & 1.436 & 0.371 & 0.800 \\
$\mathbf{T}^\mathbb{R}$: +$\mathbf{L}_\mathrm{pts}$ & \textbf{0.118} & \textbf{0.242} & 0.609 & 1.507 & 0.376 & 0.841 \\
\bottomrule
\bottomrule
\end{tabular}
\label{tab:localization_nuscene}
\end{table}



\vspace{-5pt}
\section{Conclusion}
\label{sec:conclusion}

We presented RTMap, an end-to-end framework for real-time crowdsourcing mapping. Integrating prior-aided localization, change detection, and multi-agent crowdsourcing. Its functionality includes: (1) a multi-task architecture deployed onboard for concurrently performing localization, map change detection, vectorized HD mapping, reaching real-time performance on robust HD map servicing, (2) uncertainty-aware modeling, which probabilistically represents element uncertainties to improve localization accuracy and map quality, and (3) crowdsourced mapping, enabling continuous global map refinement via collaborative multi-agent multi-traversal data. Experiments on TbV and nuScenes datasets have demonstrated RTMap’s performance for all these tasks, with dynamic updates ensuring temporal consistency. By bridging onboard perception and asynchronous cloud computing, RTMap advances safe, adaptive autonomous driving solutions to maintain and leverage a self-evolutional memory.


\textbf{Future work.} Our future directions include incorporating more sensors (e.g., LiDAR) to form a multi-modal sensor fusion and uploading auxiliary but highly compact data for better probabilistic modeling on the cloud. Expanding validation to unstructured terrains and more complex urban environments will better establish the framework's generalizability. Besides, crowdsourcing enables a maximum-a-posteriori policy to vote out temporary results gradually in our current version. In the future, our backbone supports adding occlusion detection head as an auxiliary task to distinguish between permanent structural changes and transient obstructions.

\vspace{-5pt}

\section*{Acknowledgements}

We thank the reviewers for the valuable discussions. We respect our past and present colleagues for their unwavering dedication to pursuing innovative light-weight map solutions, whose contributions have profoundly initiated this work. This research was supported by the Zhejiang Provincial Natural Science Foundation of China under Grant No. LD24F030001.

{
\small
\bibliographystyle{ieeenat_fullname}
\bibliography{main}
}
\end{document}